# Learning Earthquake Wave Arrival Time Picking from Labels with Inaccuracies

Sen Li[1*], Xu Yang[2], S. Mostafa Mousavi[1], Anye Cao[3], Keting Fan[2], Yaoqi Liu[3], Changbin Wang[4], and Qiang Niu[2]

[1] Department of Earth and Planetary Sciences, Harvard University, Cambridge, MA, USA.
[2] School of Computer Science and Technology, China University of Mining and Technology, Xuzhou, Jiangsu, China
[3] School of Mines, China University of Mining and Technology, Xuzhou, Jiangsu, China
[4] State Key Laboratory of Coal Exploration and Intelligent Mining, China University of Mining and Technology, Xuzhou, Jiangsu, China
*Corresponding author. Email: senli@fas.harvard.edu

## Abstract

Inaccurately labeled training data, or “label noise”, poses a significant threat to the integrity of supervised machine learning models. This corruption directly degrades performance by teaching the model erroneous mappings between features and labels, which leads to poor generalization and reduced accuracy on properly labeled validation and test data. Current seismological applications mainly rely on large-scale training sets or data augmentation to reduce the label-noise impact, which can be labor-intensive and costly. Here, we introduce a Label Noise-Contrastive Robust Learning (LaNCoR) approach that can effectively handle noisy labels in seismic signal processing tasks, without requiring large-scale training datasets. In this approach, the input waveform feature and label representation distributions are aligned in the feature space to correct mislabeling and reduce its impact on the training process. We present LaNCoR’s performance on the task of P-phase arrival-time picking of real microseismic data using two baseline models and training approaches. Our results indicate that LaNCoR can improve performance by up to 28.8% across performance metrics. This approach holds great promise for model training in seismology and geosciences.

# Introduction

Microseismic monitoring (Maxwell, 2014) is the high-precision detection, location, and characterization of low-magnitude seismic events (typically $M < 2$). Such low-magnitude events can arise from natural processes as well as from human activities, while dense instrumentation is especially common in induced-seismicity settings (Ellsworth 2013). By deploying dense arrays of seismometers or geophones, either on the surface or within boreholes, this technique captures weak seismic signals to map subsurface processes and assess geohazards in near-real-time. In industry, it is an indispensable tool for subsurface energy development, including hydraulic fracturing, geothermal energy, carbon sequestration, and mining monitoring. It provides critical data to map fracture network evolution, estimate stimulated reservoir volume (SRV), monitor reservoir integrity, monitor rock structural changes, assess rock instability, and mitigate dynamic hazards (Galvin 2016; Mousavi et al., 2016; Zhang et al. 2017; Cai and Kaiser 2018; Duan et al. 2024). In seismological research, microseismic datasets offer an unparalleled view of fault mechanics, illuminating active fault structures, characterizing in-situ stress fields, and providing fundamental insights into earthquake nucleation physics and fluid-rock interactions (Stevenson 1976; Takanami and Kitagawa 1991).

Deep learning has become a transformative tool in seismology, enabling the analysis of massive, complex waveform datasets by leveraging neural networks to extract subtle patterns more effectively than traditional methods (Mousavi and Beroza, 2022; 2023). This has revolutionized key domains, starting with earthquake detection and phase picking, where models like PhaseNet (Zhu and Beroza 2019) and EQTransformer (Mousavi et al. 2020) are creating catalogs an order of magnitude more complete, thus revealing new fault structures and earthquake physics (e.g. Park et al., 2020; Mohammadigheymasi et al., 2023; Yoon et al., 2023; Zhao et al., 2023). Despite the significant progress in developing deep-learning phase pickers (e.g. Feng et al. 2023; Xu et al. 2022; Chen and Li 2022), important challenges remain, including sensitivity to domain-knowledge-based preprocessing choices, the separation between detection confidence and timing probability, domain shifts between training and deployment data, and the quality of phase-picking labels (Lomax et al. 2024; Park et al. 2025)..

One of these challenges is model training on imperfectly labeled datasets that include inaccurate phase picks or “label noise” (Aguilar Suarez and Beroza, 2024; 2025). Carefully assembled seismic datasets such as STEAD and CREW include extensive quality-control procedures, yet label quality remains a central concern when models are trained or evaluated on human-picked or automatically picked arrivals (Mousavi et al. 2019; Aguilar Suarez and Beroza, 2024). Label noise is often introduced by analysts due to the subjectivity of seismic data processing/interpretation tasks and/or instrument timing errors, which can pose a threat to the validity of supervised machine learning models. These imperfectly labeled training data directly compromise the learning process, leading to a reduction in model accuracy and poor generalization to new data by teaching incorrect feature-label mappings (Song et al. 2023). Models trained on noisy data are also prone to overfitting, as they may memorize the erroneous examples rather than capturing the underlying data distribution. This issue extends to evaluation, where noisy test sets yield skewed and unreliable performance metrics, obscuring the model's true capabilities (Northcutt et al. 2021). The impact of label noise is particularly severe when it is systematic (e.g. training on auto picks), as this causes the model to learn fundamentally incorrect rules, which can lead to biased microseismic location estimates.

Although traditional regularization techniques, such as data augmentation and weight decay, have been widely adopted to mitigate overfitting, they are insufficient to address the adverse effects of noisy labels (Algan and Ulusoy 2021; Cordeiro & Carneiro 2020). Consequently, robust learning from noisy labels has become a critical research area in modern machine learning. To tackle this challenge, numerous methods have been proposed in the machine learning community, including data cleaning (Zheng et al. 2021), optimization of loss functions (Zhang and Sabuncu 2018; Wang et al. 2019), and probabilistic approaches (Liu and Tao 2015; Liu et al. 2017), aiming to learn from noisy labels and improve robustness. However, these methods often suffer from issues such as sample mis-cleaning, limited applicability, and exacerbated overfitting. As a result, despite some progress in learning with noisy labels, achieving satisfactory results with limited high-quality samples remains challenging.

In this study, we propose a novel approach, Label Noise-Contrastive Robust Learning (LaNCoR), to effectively handle label noise in seismic training datasets, particularly under small-scale training regimes. LaNCoR integrates robust learning strategies designed to distinguish between the true physical signal and the erratic noise inherent in the labels. By enforcing a statistical consistency between the stable physical features of the seismic waveform and the target labels, LaNCoR reduces the influence of inaccurately labeled samples. Instead of blindly memorizing potentially incorrect picks, the model learns to 'align' the waveform representations with the label information, correcting for inconsistencies during the training process. Our approach is inspired by the concepts of fuzzy learning (Geng 2016), and Label Distribution Learning (LDL) which are utilized for handling data ambiguity. In LDL, a label distribution is assigned to each instance and learns a mapping from instances to label distributions (Zhao, An, et al. 2023). LaNCoR first maps input seismograms and their associated labels into separate latent spaces. It then leverages the latent consistency between distributions of the feature and label representations to establish a reasonable relationship between the feature representation distribution of microseismic seismograms and their associated label representation distribution. Its objective is to learn the label errors present in fuzzy data from both waveform feature and label representations, thereby obtaining an effective representation of label errors. LaNCoR aligns feature and label representation distributions across multiple representation spaces to mitigate the negative effects of label noise, enabling high-performance deep-learning models.

In the following sections, we will first explain the methodology and then demonstrate its performance on real data using baseline models and training approaches. The proposed approach offers a promising direction for enhancing the robustness of deep-learning models in seismic signal analysis.

# Conceptual Framework

To bridge the gap between the seismological problem of noisy phase-picking labels and the deep learning solution proposed here, we first outline the conceptual basis of our approach without delving into mathematical intricacies.

## The Challenge: Distributional Misalignment

In supervised deep learning, a model learns to predict arrival times by finding a mapping between the input data (the seismic waveform) and the target (the analyst pick). In an ideal dataset, the waveform feature (e.g., the sharp amplitude jump of a P-wave) aligns perfectly with the label (the recorded time). However, when labels contain errors—whether due to human subjectivity, fatigue, or systematic timing issues—this relationship breaks down. We can visualize this as a "distributional misalignment." Imagine the waveform features occupying one space and the labels occupying another. If the labels are accurate, these two distributions overlap consistently. When noise is introduced, the label distribution shifts away from the true waveform-feature distribution. A standard model tries to chase these shifting, incorrect labels, forcing it to memorize errors rather than learning the physical characteristics of a seismic phase.

## The Strategy: Dual-Space Alignment

LaNCoR (Label Noise-Contrastive Robust Learning) addresses this by treating the waveform and the label as two different "views" of the same physical event. Instead of blindly trusting the provided label, the model projects both the waveform and the label into a shared, abstract "latent space" (a compressed mathematical representation of the data). The core intuition is **consistency**. Even if a label is incorrect, the waveform itself retains the true physical information of the arrival. LaNCoR leverages this internal consistency to "align" the data. It forces the representation of the label to look like the representation of the waveform. If they disagree (due to noise), the model uses the robust features of the waveform to pull the label's representation back into alignment (see Fig.1).

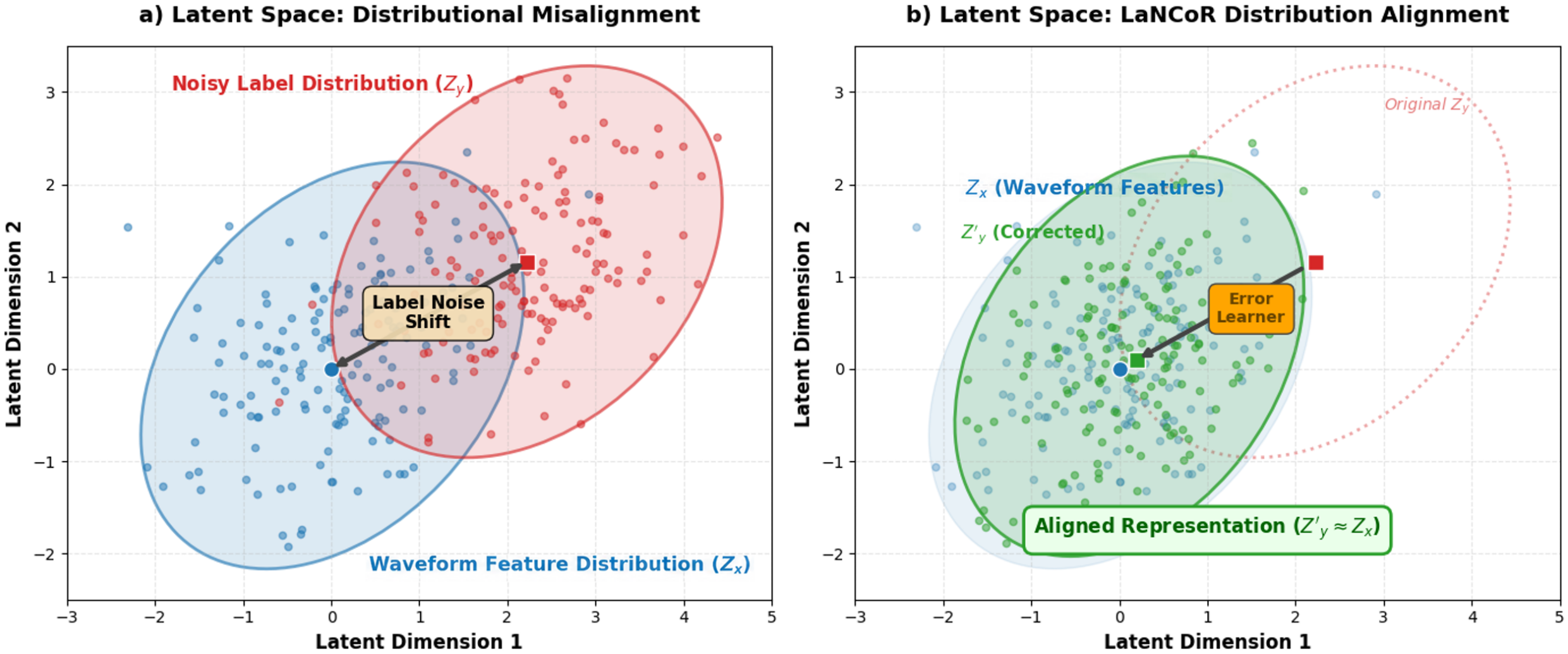

*Figure 1: Conceptual illustration of the LaNCoR strategy. (a) The Problem: In real-world seismic catalogs, manual picks (labels) often deviate from the true phase of arrival due to human error or noise. In the model's waveform-feature space, this creates a "misalignment" where the distribution of provided labels ($Z_y$) is shifted away from the distribution of true waveform features ($Z_x$), causing standard models to learn incorrect patterns. (b) The Solution: LaNCoR treats the waveform and the label as two views of the same event and estimates this shift and "pulls" the label representation back into alignment with the waveform's physical features. This forces the*

*model to learn from the consistent, physical signal rather than the erratic noise, ensuring robust picking performance.*

## The Mechanism: Learning the Error

To achieve this alignment practically, we introduce a module called the **Error Learner**. This component acts as a correction mechanism. It compares the waveform's features against the provided label's features. By analyzing the discrepancy between them, it estimates the "shift" caused by the noise. Essentially, the model learns to ask: *"Given that this waveform clearly looks like a sharp P-arrival, how far off is the provided label?"* It then corrects the label's representation internally before using it to train the final predictor. This process allows LaNCoR to filter out the noise and learn from the underlying physical signal, ensuring robust performance even when the training catalog is unreliable

# Methods

## Overall Workflow

We denote an individual seismogram instance as $\mathbf{x}_i$, and its corresponding arrival time label as $y_i$. The $j$th sampling point of $\mathbf{x}_i$ is denoted as $x_i^{(j)}$. The description degree of $y_i$ with respect to $\mathbf{x}_i$ is denoted as $\mathbf{g}_i$, where $\mathbf{g}_i = \left[g_i^{(1)}, g_i^{(2)}, \cdots, g_i^{(l)}\right]$, and $g_i^{(j)} \in [0,1]$ represents the description degree of $x_i^{(j)}$, where $l$ denotes the length of the seismogram. In this study, $l = 6000$ . For the training set $D_\delta = \{(\mathbf{x}_i, y_i)\}_{i=1}^N$ with errors, assuming an error $\delta_m$ exists in the labels, the relationship between the label $y_i$ and the true arrival time label $y_i^*$ is $y_i = y_i^* + \delta_m$. The input space is denoted as $\mathcal{X} = \mathbb{R}^l$, and the label space as $\mathcal{Y} = 1,2, \cdots, $l, where $x_i \in \mathcal{X}$ and $y_i \in \mathcal{Y}$. Since $y_i^*$ cannot be directly observed, the task is to learn the mapping relationship from the instance set $\{\mathbf{x}_i\}_{i=1}^N$ to the corresponding noisy label set $\{y_i\}_{i=1}^N$, aiming for higher performance on the accurate label dataset $D_r = \{(\mathbf{x}_i, y_i^*)\}_{i=1}^M$.

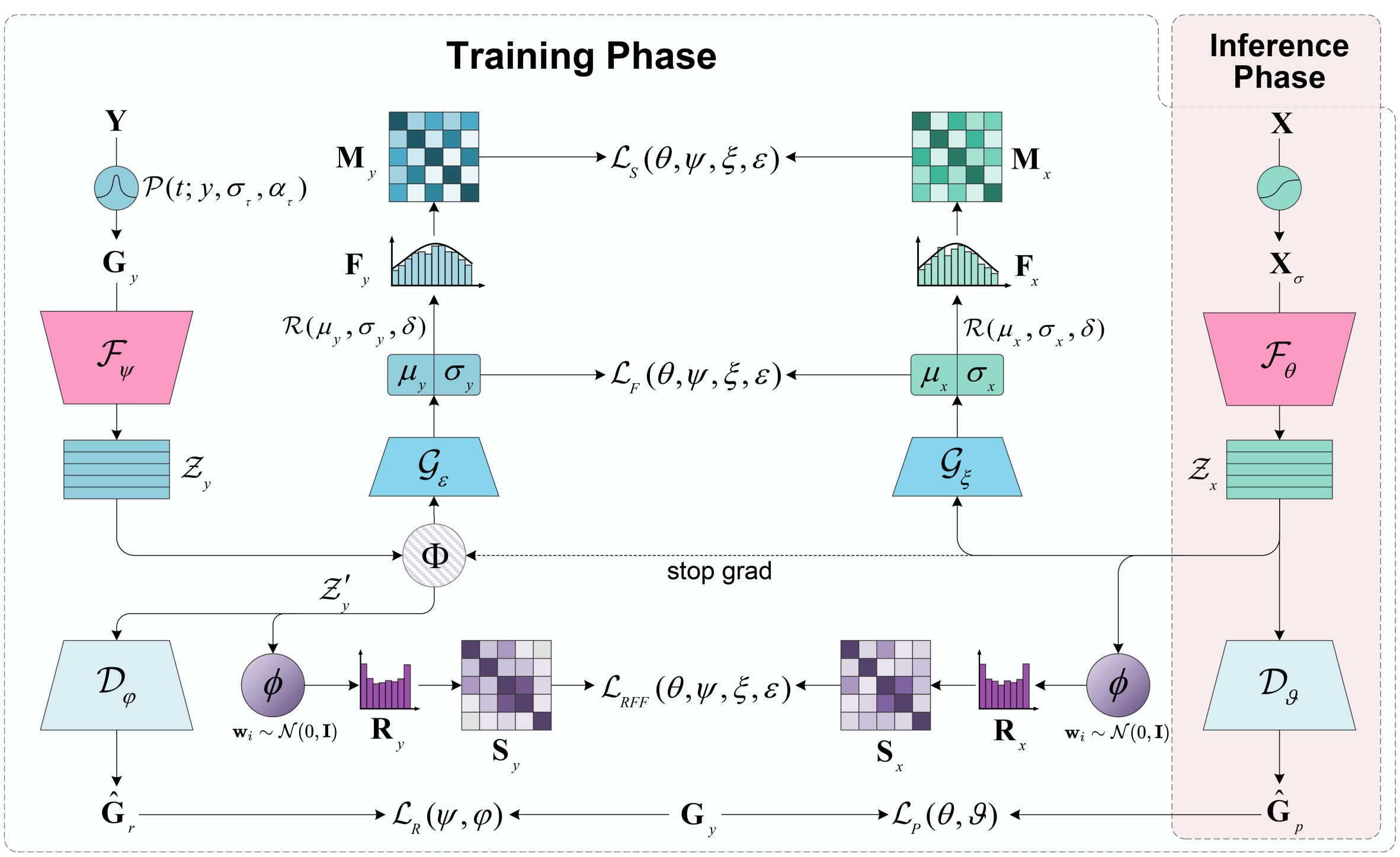

*Figure 2: The architecture of the LaNCoR framework. The model treats the input waveform (X) and the noisy label (Y) as two views of the same event. (Left/Training Phase) The input waveform and label are mapped into a shared latent space ($\mathcal{Z}_x$, $\mathcal{Z}_y$). A key innovation is the Error Learner ($\Phi$), which estimates the misalignment between the waveform features and the label features to produce a corrected label representation ($\mathcal{Z}_y{}'$). The model then minimizes the difference between these representations using Contrastive Learning (similarity matrices* **M** *and* **S***) and KL-divergence. (Right/Inference Phase) Only the trained Feature Encoder ($\mathcal{F}_\theta$) and Decoder ($\mathcal{D}_\theta$) are used to predict the arrival time ($\hat{y}$) from new waveforms. $\theta$, $\psi$, $\xi$, $\epsilon$, $\eta$, $\varphi$, and $\vartheta$ are learnable parameters; $\mathcal{P}(t; y, \sigma_\tau, \alpha_\tau)$ represents a mapping function that maps the original label to a Gaussian distribution $\mathcal{N}(y, \sigma_\tau^2)$ scaled by a scaling factor $\alpha_\tau$;* $\mathbf{G}_y$ *represents the ground truth;* $\mathbf{X}_\sigma$ *represents the normalized waveform; $\mathcal{Z}_y$ and $\mathcal{Z}_x$ represent the feature of the labels and of the waveforms, respectively; $\mathcal{R}(\mu_y, \sigma_y, \delta)$ and $\mathcal{R}(\mu_x, \sigma_x, \delta)$ represent the reparameterization trick, introducing a variable $\delta \sim \mathcal{N}(0,1)$.*

Different from traditional end-to-end learning, LaNCoR employs a novel training mechanism (Fig. 2) that can simultaneously learn the representations of input seismograms and output labels, capturing the feature representation distribution relationship between sample pairs. In fact, input features and target label distributions are two descriptions of the same instance, and their representation spaces have a natural latent consistency (Zhao, An, et al. 2023). The error in arrival time labels causes a shift in the label distribution relative to the feature representation distribution of seismograms. The goal of LaNCoR is to capture this distribution shift during training and align the label distribution and feature representation distribution of seismograms in a latent feature space, reducing the negative impact of individual inaccurate labels on model building. Specifically, it first, normalizes the original seismogram $\mathbf{x}_i$ to obtain the normalized seismogram $\mathbf{x}_i^{\sigma}$, and preprocesses the arrival time labels $y_i$ to obtain the distribution $g_i$ along the time dimension. Then, using two batch-wise mapping functions $\mathcal{F}_\theta$ and $\mathcal{F}_\psi$, it maps $\mathbf{X}_\sigma$ (a batch of $x_i^{\sigma}$) and $\mathbf{G}_y$ (a batch of $\mathbf{g}_i$) to the latent feature space, obtaining $\mathcal{Z}_x = \mathcal{F}_\theta(\mathbf{X}_\sigma)$ and $\mathcal{Z}_y = \mathcal{F}_\psi(\mathbf{G}_y)$. Then, it uses the label error learner $\Phi_\eta$ to estimate the shift of label representation distribution ($\mathcal{Z}_y$) from feature representation distribution ($\mathcal{Z}_x$) and corrects it ($\mathcal{Z}_y'$).

A key step in LaNCoR is to align the distribution relationship between $\mathcal{Z}_x$ and $\mathcal{Z}_y'$, assuming that the representation distributions of features and labels both follow Gaussian distributions. Two mapping functions $\mathcal{G}_\xi$ and $\mathcal{G}_\epsilon$ are used to map $\mathcal{Z}_x$ and $\mathcal{Z}_y'$ to $\mathcal{N}(\mu_x, \sigma_x^2)$ and $\mathcal{N}(\mu_y, \sigma_y^2)$ distributions, respectively. Meanwhile, the Fourier feature mapping $\phi$ (Tancik et al., 2020) is used to map $\mathcal{Z}_x$ and $\mathcal{Z}_y'$ to the same Fourier feature space, obtaining the low-dimensional representations $\mathbf{R}_x$ and $\mathbf{R}_y$. These are then used to calculate similarity matrices for the $\mathcal{L}_{RFF}$ contrastive loss, further aligning the feature and label representation distributions. Finally, use two mapping functions $\mathcal{D}_\vartheta$ and $\mathcal{D}_\varphi$ to decode $\mathcal{Z}_x$ and $\mathcal{Z}_y'$ to label distributions $\widehat{\mathbf{G}}_r$ and $\widehat{\mathbf{G}}_p$, respectively. During inference and deployment, the model only needs to include the feature encoder $\mathcal{F}_\theta$ and the decoder $\mathcal{D}_\vartheta$ to compute the batch-wise prediction $y$, that is:

$$\hat{y} = \text{argmax}_y\ \mathcal{D}_\vartheta(\mathcal{F}_\theta(\mathbf{X}_\sigma)).$$

To achieve multiple learning objectives, we design multiple loss functions. Reconstruction loss, $\mathcal{L}_R$, is used to learn reliable label representations, which can be computed as:

$$\mathcal{L}_R(\psi, \varphi) = \sum_{i=1}^{N} \sum_{j=1}^{l} g_i^{(j)} ln \frac{g_i^{(j)}}{D_\varphi(\mathcal{Z}_y')_i^{(j)}}.$$

$\mathcal{L}_P$ is used to learn the mapping from latent feature representations to predicted label distributions, which can be computed as:

$$\mathcal{L}_P(\theta, \vartheta) = \sum_{i=1}^{N} \sum_{j=1}^{l} g_i^{(j)} ln \frac{g_i^{(j)}}{D_\vartheta(\mathcal{Z}_x)_i^{(j)}}.$$

The final optimization objective of LaNCoR is as follows:

$$\min_{\theta,\psi,\xi,\epsilon,\vartheta,\varphi} \lambda_1 \mathcal{L}_S(\theta,\psi,\xi,\epsilon,\eta) + \lambda_2 \mathcal{L}_F(\theta,\psi,\xi,\epsilon,\eta) + \lambda_3 \mathcal{L}_{RFF}(\theta,\psi,\xi,\epsilon,\eta) + \lambda_4 \mathcal{L}_R(\psi,\varphi,\eta) + \lambda_5 \mathcal{L}_P(\theta,\vartheta),$$

where $\mathcal{L}_S$ aligns the reparameterized feature and corrected label representations in the latent feature space, while $\mathcal{L}_{RFF}$ aligns the representations in the random Fourier feature space. $\mathcal{L}_F$ is used to learn the feature representation mapping in the latent feature space, and $\{\lambda_i\}_{i=1}^{5}$ represents the loss balancing coefficients.

# Feature Alignment

Seismic data possesses an inherent duality: the physical waveform contains the true signal, while the label provides a human (or machine) interpretation of that signal. In an ideal scenario, these two representations are perfectly consistent. However, label noise breaks this consistency, creating a "distributional shift" where the label representation drifts away from the true feature representation. The goal of Feature Alignment is to force these two distributions back into agreement. By mathematically constraining the label representation to match the structure of the waveform representation, we filter out the erratic noise inherent in the labels.

The feature alignment is performed through a set of interconnected steps and using several keys that together make up the full feature alignment. More specifically, $\mathcal{L}_F$ realizes the distribution-level alignment by minimizing the KL divergence. $\mathcal{L}_s$ and $\mathcal{L}_{RFF}$ perform alignment in different feature spaces by lerveraging batch-wise contrastive learning.

## Error Learner

The primary obstacle to robust training is the unknown error $\delta_m$ embedded in the provided labels. If the model naively trusts the input label $\mathcal{Z}_y$, it will learn incorrect patterns. The Error Learner ($\Phi_\eta$) acts as a correction mechanism. It compares the questionable label representation $\mathcal{Z}_y$ against the robust waveform representation $\mathcal{Z}_x$. By analyzing the discrepancy between them (using attention mechanisms), it estimates the "shift" caused by the noise and produces a corrected label representation of $\mathcal{Z}_y{}'$. This allows the network to train targets that are mathematically closer to the ground truth than the original noisy labels.

The Error Learner ($\Phi_\eta$) adapts the multi-head self-attention mechanism（Vaswani et al. 2017) to learn and correct for the distribution misalignment between the feature representation $\mathcal{Z}_x$ of seismograms and the label representation $\mathcal{Z}_y$, counteracting label noise to some extent. This enables the corrected label representation $\mathcal{Z}_y'$ and the seismogram features representation to capture the correlated information among them better and improve the robustness and generalization of the model.

For the feature representation mapping functions $\mathcal{F}_\theta$ and $\mathcal{F}_\psi$, we can obtain the initial representations $\mathcal{Z}_x$ and $\mathcal{Z}_y$. The computation of the $\mathcal{Z}_y'$ can be expressed as:

$$\mathcal{Z}_y' = \Phi_\eta(\mathcal{Z}_x, \mathcal{Z}_y),$$

where the error learner function $\Phi_\eta(\cdot,\cdot)$ can be computed as:

$$\Phi_\eta(\mathcal{Z}_x, \mathcal{Z}_y) = \mathbf{H}^\top \mathbf{W}^{(H)} + \mathbf{b}^{(H)},$$

$$\mathbf{H} = \mathrm{C}(\mathbf{H}_1, \mathbf{H}_2, \cdots, \mathbf{H}_i),$$

$$\mathbf{H}_i = \mathrm{Softmax}\left(\frac{\mathbf{Q}_i^\top \mathbf{K}_i}{\sqrt{d_h}}\right)\mathbf{V}_i^\top,$$

$$\mathbf{Q}_i = \Omega(\mathcal{Z}_x)^\top \mathbf{W}_i^{(Q)} + \mathbf{b}_i^{(Q)},$$

$$\mathbf{K}_i = \mathrm{BN}\big(\mathrm{Pool}(\mathcal{Z}_y)\big)^\top \mathbf{W}_i^{(K)} + \mathbf{b}_i^{(K)},$$

$$\mathbf{V}_i = \mathrm{BN}\big(\mathrm{Pool}(\mathcal{Z}_y)\big)^\top \mathbf{W}_i^{(V)} + \mathbf{b}_i^{(V)},$$

where $\mathsf{C}(\cdot)$ is the concatenation operation; $\Omega(\cdot)$ represents the operation of detaching from the current computational graph; $\mathbf{H}_i$ represents the of the $i$th attention head output; $\mathbf{W}^{(H)} \in \mathbb{R}^{d_H \times d_r}$ and $\mathbf{b}^{(H)} \in \mathbb{R}^{d_r}$ are the learnable parameter matrices and bias parameter vectors, respectively; $\mathbf{Q}_i$, $\mathbf{K}_i$ and $\mathbf{V}_i$ respectively represent the query, key and value vectors of the $i$th attention head; $\mathbf{W}_i^{(Q)} \in \mathbb{R}^{d_r \times d_h}$, $\mathbf{W}_i^{(K)} \in \mathbb{R}^{d_r \times d_h}$ and $\mathbf{W}_i^{(V)} \in \mathbb{R}^{d_r \times d_h}$ are the learnable parameter matrices of the $i$th attention head respectively; $\mathbf{b}_i^{(Q)} \in \mathbb{R}^{d_h}$, $\mathbf{b}_i^{(K)} \in \mathbb{R}^{d_h}$ and $\mathbf{b}_i^{(V)} \in \mathbb{R}^{d_h}$ are the learnable bias parameter vectors; $\mathrm{BN}(\mathbf{x}) = \frac{\mathbf{x} - \mathbb{E}(\mathbf{x})}{\sqrt{\mathrm{Var}(\mathbf{x}) + \epsilon}} \cdot \gamma + \beta$, where $\gamma$ and $\beta$ are learnable affine transformation parameters, and $\epsilon$ is a small constant. We perform average pooling $\mathrm{Pool}(\mathcal{Z}_y)$ to reduce computational costs. The gradient flow from the Error Learner back to the seismic wave feature representation ($\mathcal{Z}_x$) is truncated (or detached) to maintain the integrity of the feature encoder $\mathcal{F}_\theta$. This prevents the noisy label space from corrupting the core feature learning process, which is a key part of the model's stability

## Reparameterization

Standard neural networks typically map inputs to fixed, deterministic points in feature space. However, when data is noisy, deterministic mapping can lead to "overconfidence," where the model memorizes specific noisy examples. To prevent this, we employ Reparameterization to treat the features and labels not as fixed points, but as probability distributions (Gaussian clouds) with a mean ($\mu$) and variance ($\sigma$). This introduces a controlled degree of uncertainty during training, preventing the model from overfitting to exact (but potentially wrong) values and smoothing the learning landscape.

LaNCoR uses common methods, such as variational inference (Blei et al., 2017), to control the distributions of feature and label representations.

Specifically, we define a prior distribution and sample the seismogram feature representation $\mathcal{Z}_x$ using a reparameterization trick (Rezende et al. 2014), i.e., $\mathbf{F}_x = \mu_x + \sigma_x \delta$, where $\delta \sim \mathcal{N}(0, I_{N \times d_z})$, and both $\mu_x$ and $\sigma_x$ are computed by $\mathcal{G}_\xi(Z_x)$, resulting in $\mathbf{F}_x \in \mathbb{R}^{N \times d_z}$. Similarly, we reparameterize $\mathcal{Z}'_y$ as $\mathbf{F}_y = \mu_y + \sigma_y \delta$, where both $\mu_y$ and $\sigma_y$ are computed by $\mathcal{G}_\epsilon(\mathcal{Z}'_y)$, resulting in $\mathbf{F}_y \in \mathbb{R}^{N \times d_z}$. By expressing these random variables as deterministic functions of the model's parameters and a separate random noise variable ($\delta$), it allows for the gradients to be computed with respect to the model parameters $\mu_x, \sigma_x, \mu_y$ , and $\sigma_y$ moving the randomness outside the model. This trick is crucial for end-to-end variational model training, and the uncertainty introduced by the reparameterization trick is beneficial for improving the generalization of the model. The objective of $\mathcal{L}_F$ is to improve the model performance by utilizing the relationship between label representation distribution and feature representation distribution (Zhao, Qi, et al. 2023), i.e., optimizing the KL divergence between $\mathcal{N}(\mu_x, \mathrm{diag}(\sigma_x^2))$ and $\mathcal{N}\big(\mu_y, \mathrm{diag}(\sigma_y^2)\big)$, which can be expressed as:

$$\mathcal{L}_F(\theta,\psi,\xi,\epsilon,\eta) = \mathcal{KL}\left(\mathcal{N}(\mu_x, \text{diag}(\sigma_x^2))||\mathcal{N}\left(\mu_y, \text{diag}(\sigma_y^2)\right)\right)$$
$$= \frac{1}{2}\sum_{i=1}^{d_\chi}\left(\log\sigma_y^{(i)2} - \log\sigma_x^{(i)2} + \omega^{(i)}\right),$$

where $\sigma^{(i)}$ and $\mu^{(i)}$ represent the variance and mean of the $i$th dimension, respectively; $\omega^{(i)} = \frac{\sigma_x^{(i)2}+\left(\mu_x^{(i)}-\mu_y^{(i)}\right)^2}{\sigma_y^{(i)2}} - 1$; $d_\chi$ denotes the dimensionality of the latent feature space.

## Random Fourier Features

Aligning distributions in a single, low-dimensional space may not capture all the complex, non-linear relationships between waveforms and labels. To ensure a more rigorous alignment, we project the data into a high-dimensional space using Random Fourier Features (RFF) (Tancik et al. 2020). This technique approximates a Gaussian kernel, allowing us to align the high-frequency details and non-linear structures of the distributions. Essentially, this ensures that the waveform and label representations match not just in their general position, but in their fine-grained structural details.

Specifically, we use the random projection of the approximate non-linear kernel method (Rahimi and Recht 2007) to achieve random feature mapping $\phi: \mathbb{R}^{d_r} \rightarrow \mathbb{R}^{d_z}$, which can be computed as:

$$\phi(\mathbf{x}) = \sqrt{\frac{1}{d_\phi}}[sin(\mathbf{x}^\top\mathbf{w}_1), \cdots, sin(\mathbf{x}^\top\mathbf{w}_{d_\phi}), cos(\mathbf{x}^\top\mathbf{w}_1), \cdots, cos(\mathbf{x}^\top\mathbf{w}_{d_\phi})]^\top,$$

where $\mathbf{w}_i \sim \mathcal{N}(0, I_{d_r})$ ; $d_\phi = \frac{d_z}{2}$ . For seismogram feature representation $\mathcal{Z}_x$ and label representation $Z_y'$, we have random features $\mathbf{R}_x \in \mathbb{R}^{N\times d_z}$ and $\mathbf{R}_y \in \mathbb{R}^{N\times d_z}$. The computation is as follows:

$$\mathbf{R}_x = \phi(\mathcal{Z}_x),$$
$$\mathbf{R}_y = \phi(\mathcal{Z}_y').$$

## Contrastive Learning

Once we have corrected the label representations and projected them into the appropriate spaces, we need a loss function to enforce the alignment. We utilize Contrastive Learning to maximize the agreement between the two "views" of the event (the waveform and the label). Unlike standard losses that look at samples individually, contrastive learning looks at the relationship between samples. It enforces that if two waveforms are similar in feature space, their corresponding labels must also be similar in label space. This global structural constraint prevents the label distribution from drifting due to random noise.

The objective of $L_S$ is to align feature representation distributions in the latent feature space, thereby mitigating errors introduced by label noise and enhancing the model performance and generalization ability. Specifically, we draw upon the principles of contrastive learning (Hadsell et al., 2006; Chen et al., 2020). However, unlike standard approaches that contrast positive and negative sample pairs from the same modality, our approach aligns the distributions across two different modalities: the instance feature space and the label space. This is achieved by constructing a similarity matrix between samples. For $\mathbf{F}_x$, we have the similarity matrix $\mathbf{M}_x$, which can be computed as:

$$M_x^{(i,j)} = S\left(\mathbf{f}_x^{(i)}, \mathbf{f}_x^{(j)}\right),$$

where $M_x^{(i,j)}$ represents the element at position $(i,j)$ of similarity matrix $\mathbf{M}_x$; $i,j \in \{1,2,\dots,N\}$; $\mathbf{f}_x^{(i)} \in \mathbb{R}^{d_z}$ represents the $i$th sample in $\mathbf{F}_x$; $\mathbf{M}_x \in \mathbb{R}^{N\times N}$; $N$ is the number of samples. $S(\cdot,\cdot)$ denotes cosine similarity, and is defined as:

$$S\left(\mathbf{f}_x^{(i)}, \mathbf{f}_x^{(j)}\right) = \frac{\mathbf{f}_x^{(i)} \cdot \mathbf{f}_x^{(j)}}{\| \mathbf{f}_x^{(i)} \| \cdot \| \mathbf{f}_x^{(j)} \|}.$$

Similarly, for label representation $F_y$, we have:

$$M_y^{(i,j)} = S\left(\mathbf{f}_y^{(i)}, \mathbf{f}_y^{(j)}\right).$$

$$S\left(\mathbf{f}_y^{(i)}, \mathbf{f}_y^{(j)}\right) = \frac{\mathbf{f}_y^{(i)} \cdot \mathbf{f}_y^{(j)}}{\| \mathbf{f}_y^{(i)} \| \cdot \| \mathbf{f}_y^{(j)} \|}.$$

Then, we use mean squared error (MSE) to minimize the difference between the two similarity matrices:

$$\mathcal{L}_S(\theta,\psi,\xi,\epsilon,\eta) = \frac{1}{N^2} \| \mathbf{M}_x - \mathbf{M}_y \|_2^2.$$

The objective of $\mathcal{L}_{RFF}$ is to align the distribution of random feature representations in the random feature space. We aim to minimize:

$$\mathcal{L}_{RFF}(\theta,\psi,\xi,\epsilon,\eta) = \frac{1}{N^2} \| \mathbf{S}_x - \mathbf{S}_y \|_2^2,$$

Here, $\mathbf{S}_x$ and $\mathbf{S}_y$ are the similarity matrixes, which can be computed as:

$$S_x^{(i,j)} = S\left(\mathbf{r}_x^{(i)}, \mathbf{r}_x^{(j)}\right),$$

$$S_y^{(i,j)} = S\left(\mathbf{r}_y^{(i)}, \mathbf{r}_y^{(j)}\right).$$

where $r_x^{(i)} \in \mathbb{R}^{d_z}$ represents the $i$th sample in $\mathbf{R}_x$.

# Data

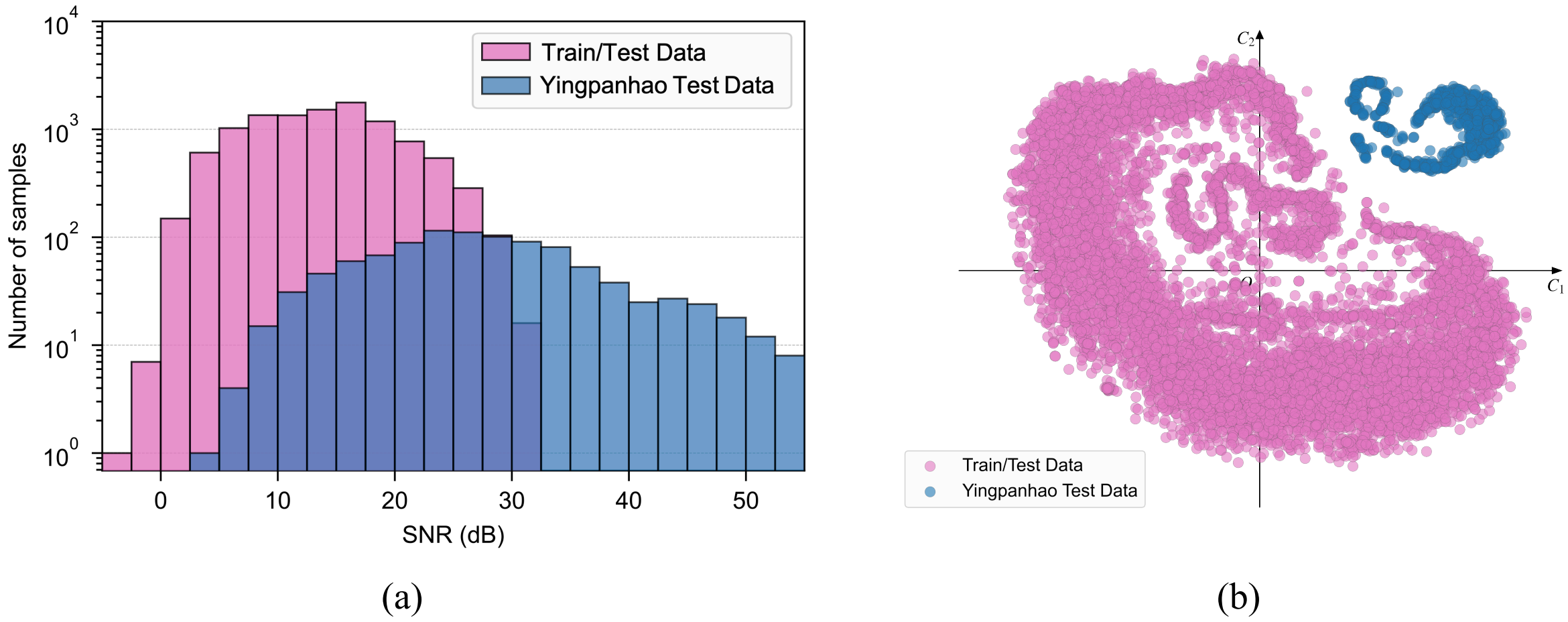

(a) (b)

*Figure 3: Characterization of the feature-distribution shift between datasets. (a) Signal-to-Noise Ratio (SNR) distributions. The histograms reveal a physical disparity: the Training/Test data (pink) is concentrated between -3.4 and 32.3 dB, while the cross-region Yingpanhao data (blue) extends to significantly higher SNRs (up to 55 dB). (b) t-SNE visualization of the learned feature space. The distinct separation between the Training cluster (pink) and the Yingpanhao cluster (blue) confirms that the physical differences shown in (a) translate into a measurable feature-distribution shift in the latent feature space, motivating the robust alignment strategy of LaNCoR.*

We used microseismic events recorded at three coal mines in China: Chaoyang Coal Mine in Shandong Province, Mengcun Coal Mine in Shaanxi Province, and Xingcun Coal Mine in Shandong Province to evaluate LaNCoR performance. The training and test datasets comprise a total of 10,685 individual waveform samples recorded at a sampling rate of 500 Hz. The SNR of waveforms ranges from -3.4 dB to 32.3 dB (Fig. 3a). We randomly divide the data into a training set (80%), validation set (10%), and test set (10%).

To evaluate the method's cross-region generalization under a measured feature-distribution shift, we utilized a test dataset from the Yingpanhao Coal Mine in Inner Mongolia. This dataset comprises 1,019 samples, recorded at a sampling rate of 500 Hz, with Signal-to-Noise Ratios (SNRs) ranging from 2.5 dB to 55.0 dB (refer to "Yingpanhao test data" in Fig. 3a). This specific dataset was exclusively reserved for testing the models' generalization capabilities and was not incorporated into the training process. The disparities in feature distribution between the two datasets are illustrated in Fig. 3b.

This approach aims to train a model with good performance under labeling error. The ideal true label $y_i^*$ is related to the available label $y_i$ as $y_i = y_i^* + \delta_m$. Since it is difficult to quantify $\delta_m$, we introduce additional Gaussian noise $\delta_s \sim \mathcal{N}(0, \sigma_s^2)$, such that $\delta_m := \delta_m + \delta_s$. This way, the training data have labels with errors, which introduces greater errors to the training set labels. Meanwhile, the test set still retains the original labels, making the test set labels relatively closer to $y_i^*$ compared to the training set, thus verifying the effectiveness of the approach even in the case of non-Gaussian error distributions. This allows us to train the model on a significantly more perturbated training set to validate the robustness of the training strategy. It is important to clarify that while we employ a zero-mean Gaussian assumption to generate synthetic training noise (ensuring controlled variable intensity), we do not assume the real-world test data follows this distribution. Real-world manual picks often contain asymmetric biases and non-Gaussian outliers. By training on Gaussian perturbations but evaluating on the unaltered human-labeled test set, we demonstrate that LaNCoR learns robust physical features that generalize well to realistic, non-Gaussian noise regimes.

## Experimental Setup

To assess LaNCoR's effectiveness, we utilized SeisT (Li et al., 2024) and PhaseNet (Zhu and Beroza, 2019) as our benchmark models. SeisT is a seismicity-focused model suitable for diverse seismic monitoring applications, including microseismic monitoring, while PhaseNet specializes in seismic phase picking. Our model training process involved specific initialization and optimization techniques. For the linear and convolutional layers in SeisT, we used a truncated normal distribution (Jinho et al. 2013), set BatchNorm weights to 1, and initialized all biases to 0. PhaseNet's weights were initialized using "kaiming uniform" (He et al. 2015). We optimized the models with the Adam optimizer (Kingma and Ba 2015), employing a cyclic learning rate scheduler (Smith 2017). The learning rate ranged from a minimum of $8 \times 10^{-5}$ to a maximum of $1 \times 10^{-3}$, with periodic adjustments to avoid local minima. Early stopping was implemented, halting training if validation performance did not improve for 30 consecutive epochs. Data augmentation, applied to the training dataset (specifically in methods labeled "Aug"), involved adding random Gaussian white noise, time shifts, introducing gaps, amplitude scaling, and generating noise. These augmentations were applied with probabilities of 0.4, 0.2, 0.4, 0.4, and 0.1, respectively. Regularization was achieved using DropPath (Huang et al. 2016) and Dropout (Srivastava et al. 2014). All models were trained within the PyTorch framework on a single NVIDIA RTX-4090 GPU.

The evaluation of these models across various tasks involved several key metrics: Precision, Recall, and the F1-score (F1). These metrics are defined as:

$$Precision = \frac{T_p}{F_p + T_p},$$

$$Recall = \frac{T_p}{F_n + T_p},$$

$$F1 = \frac{2 \times Precision \times Recall}{Precision + Recall}.$$

Here, samples with an absolute error less than the tolerance threshold ($\delta_{tol}$) are considered as true positives. $\delta_{tol}$ is ranging from 20 ms to 100 ms. $T_p$, $F_p$, and $F_n$ represent the number of true positive samples, false positive samples, and false negative samples, respectively.

# Results

## Main Results

We rigorously evaluated the performance of LaNCoR in comparison with several baseline training approaches, including supervised learning with data augmentation and LaNCoR (“Aug+LaNCoR”), supervised learning with data augmentation only (“Aug”), and supervised learning without Augmentation (“Base”). Considering the non-deterministic computations in some parts of the training process, we conducted five training and testing runs for each configuration. As shown in Table S1 and Fig. 5, the results demonstrate that LaNCoR achieves considerable performance across various levels of label noise. For the F1-score, the average improvements of SeisT with the LaNCoR-only training method over the other three baseline methods at different noise levels are 5.7%, 16.3%, and 10.1%, with maximum performance improvements of 14.7%, 28.8%, and 18.9% for Aug+LaNCoR, Aug, and Base respectively. Under the same conditions, the average improvements in precision for SeisT with LaNCoR-only over the other three baseline training approaches at different noise levels are 5.2%, 13.6%, and 9.7%, with maximum improvements of 14.7%, 28.7%, and 19.5% for Aug+LaNCoR, Aug, and Base respectively. Meanwhile, the average improvements in recall for SeisT with LaNCoR-only are 6.0%, 17.8%, and 10.3%, with maximum improvements of 14.7%, 28.8%, and 18.5% for Aug+LaNCoR, Aug, and Base respectively.

A comparative analysis of performance across various Signal-to-Noise Ratio (SNR) conditions is presented in Fig. 8 and 9. While the fundamental distribution of arrival picks is inherently shaped by the data samples themselves, variations in model performance are evident when different training methodologies are employed. The LaNCoR model consistently demonstrates notable performance enhancements across all SNR levels. This improvement is particularly critical for monitoring microseismicity and smaller earthquakes, which are inherently characterized by low SNRs. As shown in the lower SNR bins of Fig. 8 and 9 (representing data down to -3.4 dB), LaNCoR significantly outperforms the baseline methods where signal visibility is minimal. Visual confirmation of this capability is provided in Fig. 10 (a) and (b), which depicts the model’s ability to accurately recover arrival times from waveforms heavily obscured by background noise.

Beyond the SeisT model, we also conducted evaluations on PhaseNet to further assess the impact of different training approaches. When utilizing LaNCoR's training methodology, PhaseNet exhibited average F1-Score improvements over the three baseline methods at various noise levels, specifically 21.3%, 22.6%, and 17.8%. These improvements reached maximums of 54.7%, 49.6%, and 44.4%. Similarly, the average precision for PhaseNet with LaNCoR improved by 21.3%, 22.7%, and 18.3%, with peak gains of 56.1%, 49.6%, and 44.4%. Furthermore, average recall improvements were recorded at 21.4%, 22.7%, and 17.5%, achieving maximum increases of 52.1%, 50.4%, and 44.4%, respectively.

## Cross-Region Generalization Performance

To assess the generalization capabilities of the LaNCoR approach across regions, the model was applied to microseismic data from mining regions not included in its training. We use this dataset as a cross-region test with a measured feature-distribution shift, rather than assuming that geographic separation alone defines an out-of-distribution case. As detailed in Fig. 7 and Table S2,

LaNCoR consistently outperformed other methods across various values of $\sigma_s$, achieving the highest F1-Score, Precision, and Recall in most instances. While LaNCoR maintained a lead for models trained on low-noise data, this advantage was not particularly pronounced. However, as $\sigma_s$ increased, LaNCoR's superiority over the "Aug" method became more evident. For example, at = 900 ms, SeisT with LaNCoR achieved an F1-Score of 0.727, a 15.6% improvement over "Aug." As values continued to rise, "Aug" displayed considerable performance fluctuations, suggesting that data augmentation results in inconsistent model performance. Conversely, LaNCoR demonstrated more consistent outcomes, indicating its enhanced robustness to varying levels of label noise in cross-region conditions.

## Ablation Study on Error Learner (EL)

The EL simultaneously integrates information from both the label space and feature space. To validate the effectiveness of EL, we conducted ablation experiments on the validation set. As shown in Fig. 4, the results indicate that EL can effectively improve model performance. Under various levels of label noise, the average improvement of the F1-score for LaNCoR over the LaNCoR without EL at different noise levels is 4.87%, with maximum improvements of 9.55%. This demonstrates that EL captures valuable feature representation distributions from the feature space into the label space, reducing the noise level in label representations and effectively enhancing the performance of the proposed LaNCoR approach in this study.

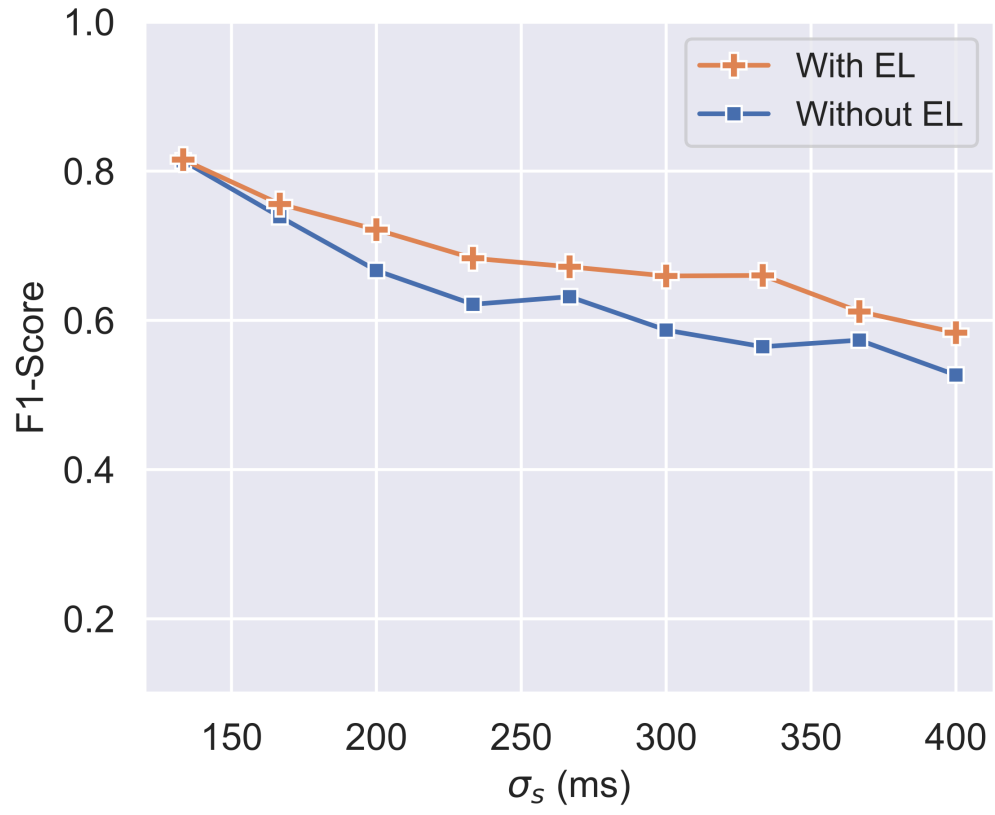

*Figure 4: The impact of the Error Learner (EL). A comparison of F1-Scores with and without the Error Learner module across increasing noise levels ($\sigma_s$). The consistent gap between the orange line (With EL) and blue line (Without EL) demonstrates that the Error Learner contributes a steady performance gain (~5% on average). This confirms that explicitly estimating the label error $\delta_m$ is crucial for the model's robustness.*

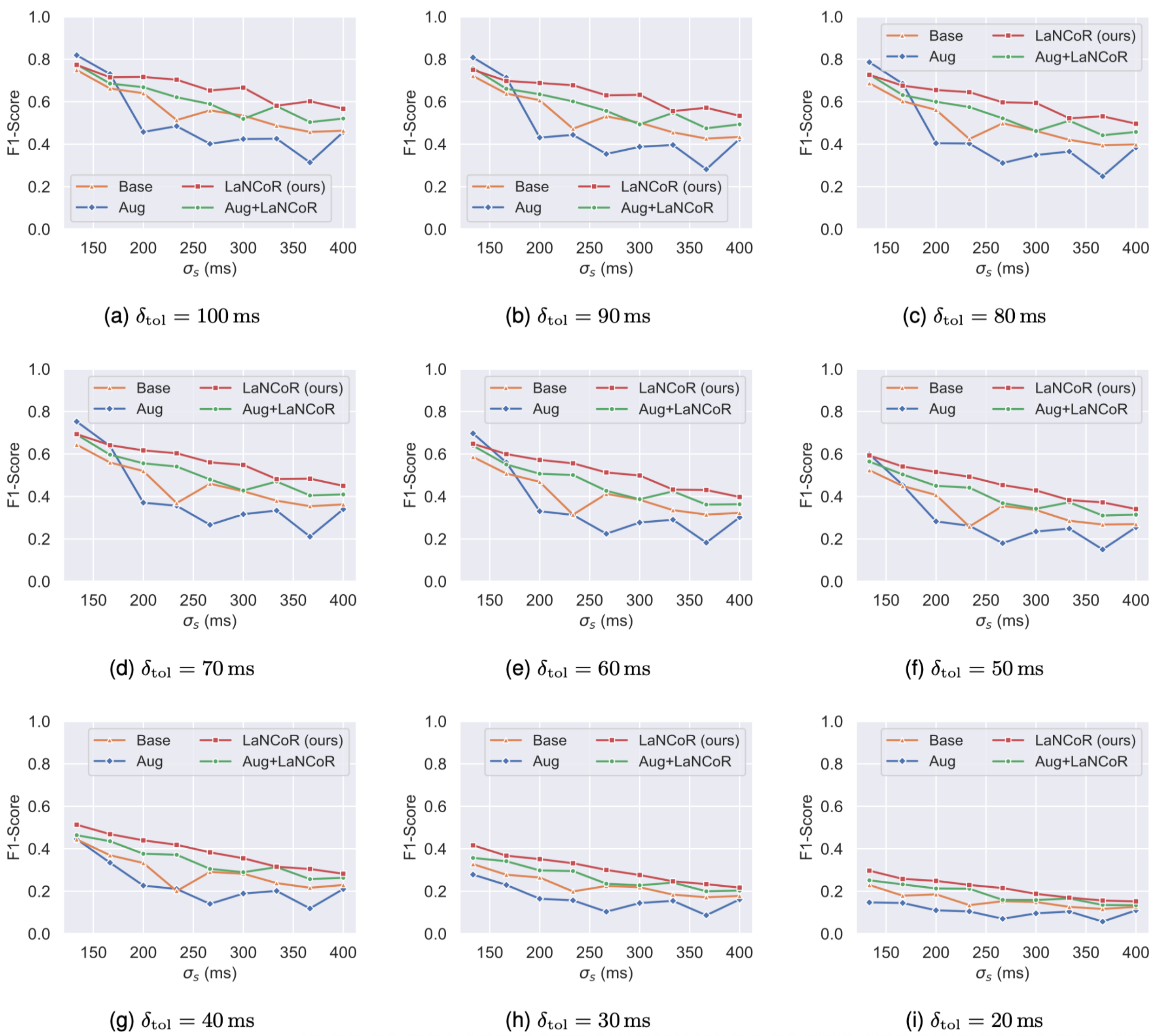

(a) $\delta_{\mathrm{tol}} = 100\,\mathrm{ms}$ (b) $\delta_{\mathrm{tol}} = 90\,\mathrm{ms}$ (c) $\delta_{\mathrm{tol}} = 80\,\mathrm{ms}$

(d) $\delta_{\mathrm{tol}} = 70\,\mathrm{ms}$ (e) $\delta_{\mathrm{tol}} = 60\,\mathrm{ms}$ (f) $\delta_{\mathrm{tol}} = 50\,\mathrm{ms}$

(g) $\delta_{\mathrm{tol}} = 40\,\mathrm{ms}$ (h) $\delta_{\mathrm{tol}} = 30\,\mathrm{ms}$ (i) $\delta_{\mathrm{tol}} = 20\,\mathrm{ms}$

*Figure 5: Performance stability under increasing noise. Comparison of SeisT trained with LaNCoR (Red) against baseline methods (Blue/Green) across varying noise levels ($\sigma_s$). Note the "crossing point" around $\sigma_s = 200$ ms. While standard augmentation (Blue) works for low noise, its performance collapses as noise increases (overfitting to errors). In contrast, LaNCoR (Red) maintains a flat, stable performance curve even at high noise levels, validating its resistance to label corruption.*

## Impact of Loss Balancing Coefficients $\lambda$

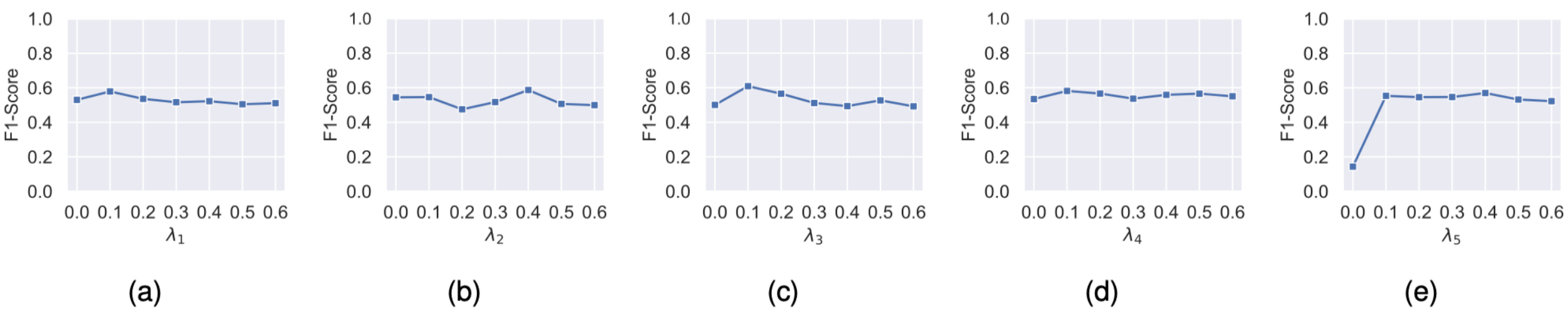

(a) (b) (c) (d) (e)

*Figure 6: Impact of loss balancing coefficients $\{\lambda_i\}_{i=1}^{5}$. Here, we used $\delta_{tol} = 100$ ms, $\sigma_s = 400$ ms, and performed evaluation on the validation set.*

(a) $\delta_{\text{tol}} = 100\,\text{ms}$

(b) $\delta_{\text{tol}} = 90\,\text{ms}$

(c) $\delta_{\text{tol}} = 80\,\text{ms}$

(d) $\delta_{\text{tol}} = 70\,\text{ms}$

(e) $\delta_{\text{tol}} = 60\,\text{ms}$

(f) $\delta_{\text{tol}} = 50\,\text{ms}$

(g) $\delta_{\text{tol}} = 40\,\text{ms}$

(h) $\delta_{\text{tol}} = 30\,\text{ms}$

(i) $\delta_{\text{tol}} = 20\,\text{ms}$

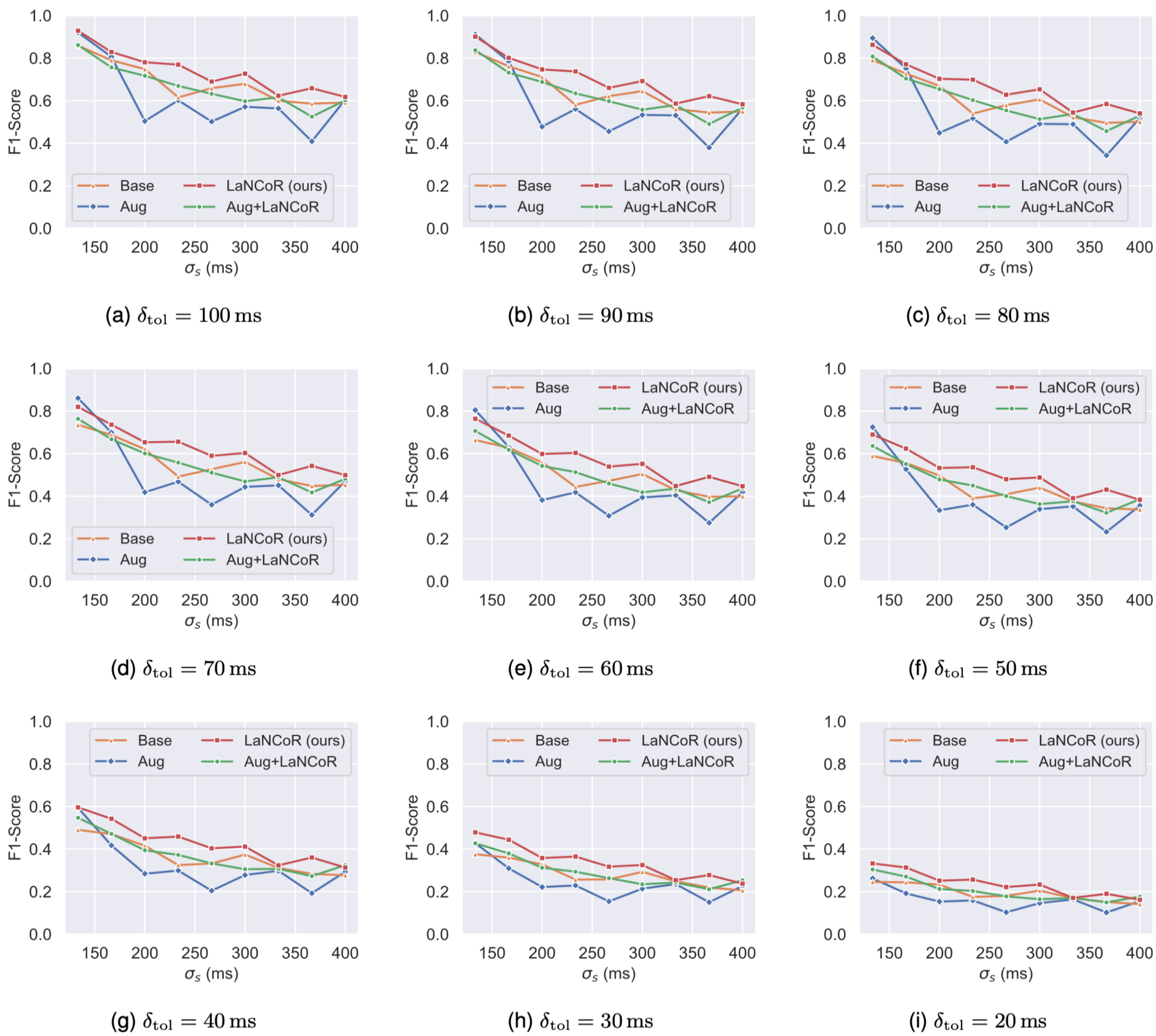

*Figure 7: Cross-region generalization under label noise. Performance on the unseen Yingpanhao dataset. LaNCoR (Red) consistently outperforms the baseline across all noise levels. This confirms*

*that LaNCoR does not just memorize training data but learns physical waveform features that remain valid in data from a different mining region with a measured feature-distribution shift, even when the training labels were highly unreliable.*

To explore the impact of the loss balancing coefficients $\{\lambda_i\}_{i=1}^{5}$, we evaluated the performance of LaNCoR with different coefficient values on the validation set. Considering the computational cost and the scale of the coefficient search space, we chose the greedy search method. The optimal values we obtained were 0.1, 0.4, 0.1, 0.1, and 0.4, respectively. The experimental results (Fig. 6) indicate that the classification loss and contrastive alignment terms are most important. While the coefficients for feature representation distribution alignment and label reconstruction losses are relatively small, they are still indispensable. This demonstrates the effectiveness of the LaNCoR architecture design.

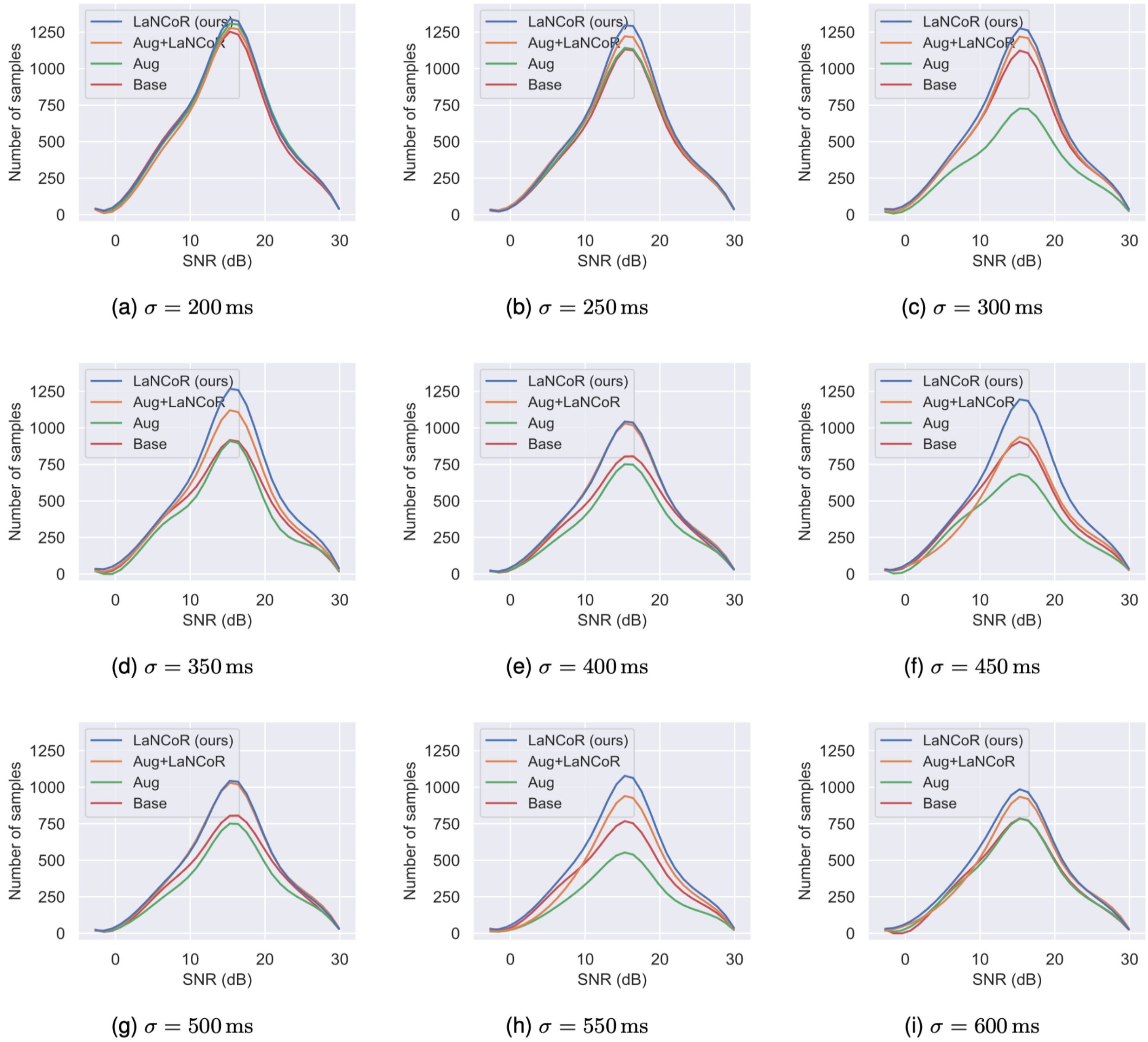

*Figure 8: Model performance of SeisT under $\delta_{tol} = 100$ ms. Here, $\sigma = 3\sigma_s$.*

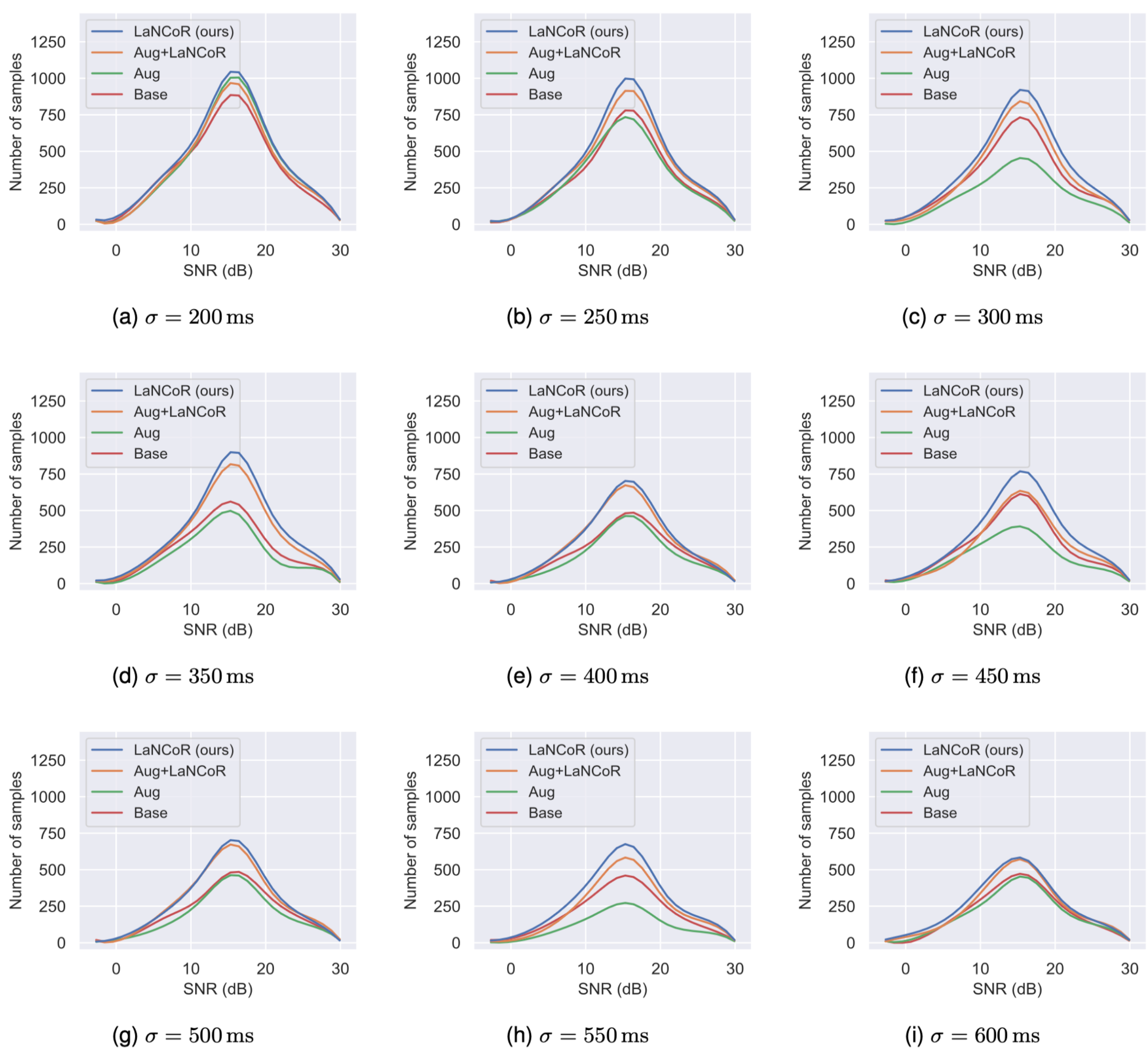

(a) $\sigma = 200\,\mathrm{ms}$ (b) $\sigma = 250\,\mathrm{ms}$ (c) $\sigma = 300\,\mathrm{ms}$

(d) $\sigma = 350\,\mathrm{ms}$ (e) $\sigma = 400\,\mathrm{ms}$ (f) $\sigma = 450\,\mathrm{ms}$

(g) $\sigma = 500\,\mathrm{ms}$ (h) $\sigma = 550\,\mathrm{ms}$ (i) $\sigma = 600\,\mathrm{ms}$

*Figure 9: Model performance of SeisT under* $\delta_{tol} = 50$ *ms. Here,* $\sigma = 3\sigma_s$.

# Qualitative Analysis

To visualize the model's behavior under different conditions, we examined representative waveform samples and their corresponding predictions (Fig. 10). Fig. 10-a and b display test samples with low Signal-to-Noise Ratios. Despite the high background noise, the model successfully identifies the P-wave onset. Fig. 10-d illustrates a case where the original manual annotation contained only a single arrival time. The LaNCoR-trained model, however, identified a second distinct P-wave arrival earlier in the trace, effectively recovering missing seismicity. Fig. 10-e and f show training samples where the provided manual labels (orange lines) were intentionally incorrect (offset from the true arrival). In both cases, the model's prediction (blue dashed line) aligns with the physical energy onset rather than the erroneous label.

# Discussion

We compared LaNCoR with several existing methods, including direct supervised learning, supervised learning with data augmentation, the combination of data augmentation and LaNCoR, and using LaNCoR only, to verify the effectiveness of LaNCoR. LaNCoR exhibits competitive or superior performance when compared to current supervised training methods, particularly across varying tolerance thresholds and noise levels. These improvements are attributed to the ability of LaNCoR to effectively capture and correct representation distribution shifts caused by label noise, aligning the latent feature representations of the seismograms with the label distributions. This alignment reduces the impact of noisy labels, thereby improving model performance.

To ensure fairness and reduce randomness, we tested the four different methods using the same hyperparameters across five different random seeds. Our experiments revealed that the level of label noise significantly influences the ranking of model performance. When $\sigma_s$ is less than 150ms, the performance differences among the methods are acceptable. With $\delta_{tol}$ less than 50ms, data augmentation results in the worst performance among all models. When the label noise level $\sigma_s$ exceeds 150ms, data augmentation no longer positively impacts model performance, showing a noticeable performance decline. At a noise level $\sigma_s$ of 200ms and above, data augmentation negatively impacts model performance, performing worse than models trained directly on raw data. We believe this is due to the combined effect of regularization from data augmentation and label noise, which increases the complexity and overall noise level of the dataset, making it difficult for the model to capture the mapping between seismic waveforms and phase labels during training, thus failing to converge to acceptable performance. Although the LaNCoR approach applied to SeisT and PhaseNet showed different final performances, LaNCoR can still train a model with higher performance than traditional training approaches like data augmentation.

It is important to note that our training strategy introduces zero-mean Gaussian noise to simulate label inaccuracies. We acknowledge that real-world analyst errors often exhibit non-Gaussian characteristics, such as a systematic bias toward picking arrival times earlier than the energy onset. However, a key strength of our evaluation is that while the training labels were perturbed with Gaussian noise, the test set retained the original human-picked labels. Consequently, the test set inherently contains the specific non-Gaussian biases and asymmetric error distributions typical of human analysts. The model’s ability to generalize from Gaussian-perturbed training data to human-labeled test data confirms that LaNCoR is robust not only to the intensity of noise but also to distributional shifts between the noise models.

LaNCoR significantly enhances the stability of model training. As illustrated in Fig. 5, methods incorporating LaNCoR exhibit smoother, more linear performance curves. This consistency supports the hypothesis that model performance is inversely related to label noise levels. In contrast, conventional methods (those not utilizing LaNCoR) show considerable performance fluctuations, with curves displaying notable oscillations (sharp drops or gains) as label noise varies, indicating their vulnerability to inaccurate labels. This instability stems from the inability of "Aug" and "Base" methods to provide robust learning guidance during training, leading to anomalous scenarios where model performance paradoxically improves with increased label noise in certain ranges. This highlights an unstable training environment. Thus, LaNCoR demonstrably fosters stable model training.

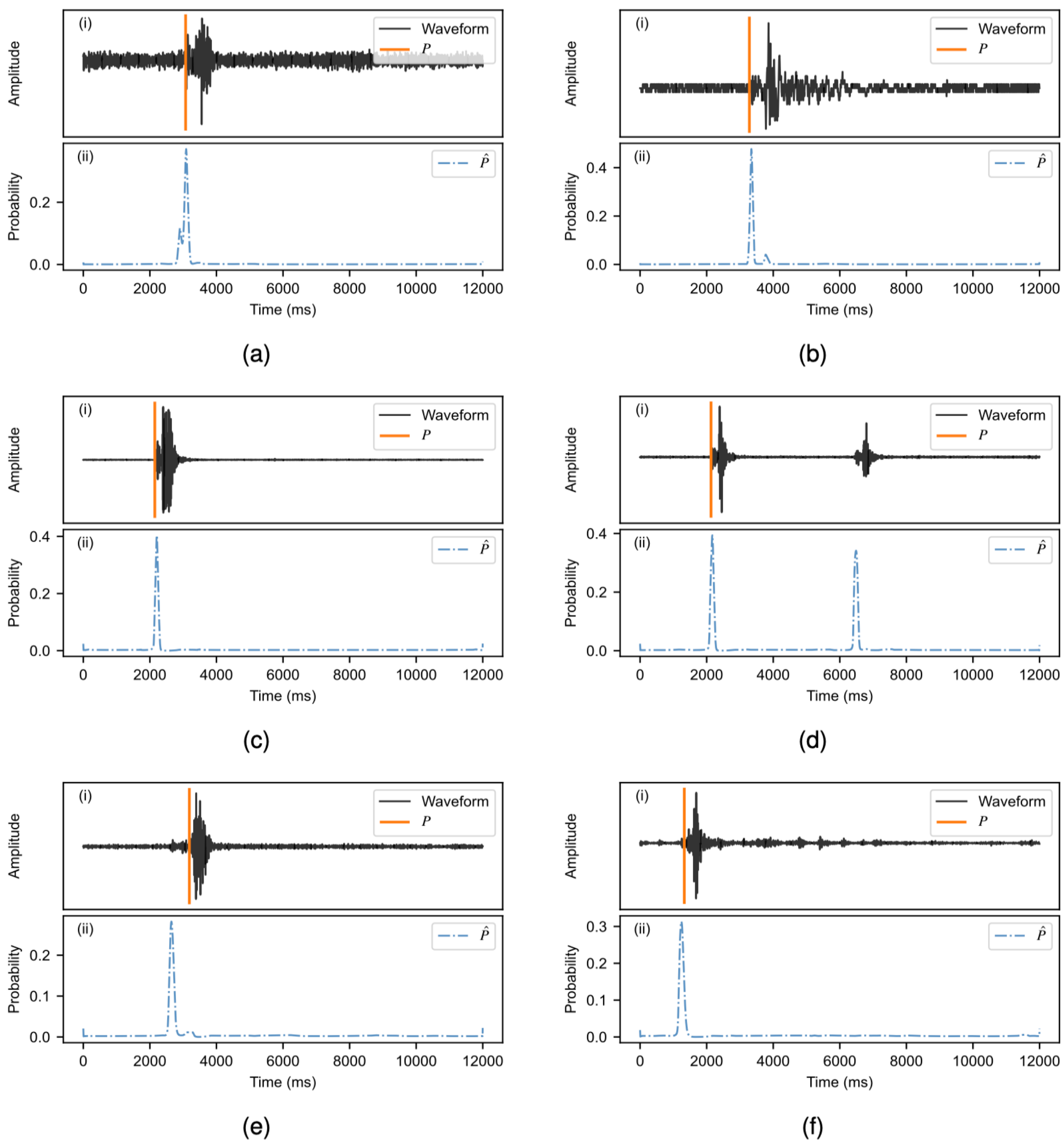

*Figure 10: Qualitative picking results. Blue dashed lines represent the predicted probability distribution; orange lines represent manual labels. (a-b) Successful recovery of arrival times in low-SNR samples, addressing the challenge of detecting small earthquakes. (d) Discovery of a missing phase: the model identifies a second P-wave arrival that was missed by the human analyst. (e-f) Resistance to bad labels: These training samples contain intentionally incorrect labels (orange line far from onset). The LaNCoR model ignores the noisy label and correctly identifies the true physical arrival, demonstrating that it has learned to prioritize signal features over erroneous supervision.*

From a data augmentation perspective, the curves with data augmentation display a certain symmetry when compared with those without. For instance, a clear symmetry is observed in the trends between the "Aug" and "Base" methods. This suggests that data augmentation substantially impacts model performance and does not consistently guarantee positive outcomes. However, the presence of LaNCoR attenuates this symmetry, implying that LaNCoR effectively mitigates the uncertainties introduced by data augmentation, thereby reinforcing its positive effect on stable model training.

Furthermore, cross-region experiments on the Yingpanhao dataset corroborate these findings. LaNCoR consistently outperforms other methods across different noise levels and maintains stability. Fig. 7 demonstrates that LaNCoR sustains relatively smooth and stable performance on the cross-region test data, unlike traditional methods, which show marked performance variability and struggle with consistent accuracy in new regions. This underscores LaNCoR's robustness in managing noisy labels and its superior adaptability.

LaNCoR enables accurate prediction of unknown P-wave arrival times and effectively handles noisy labels in the training data. The qualitative results (Fig. 10) confirm the model's resilience. The ability to distinguish true phase arrivals from background noise (Fig. 10a-b) and, crucially, to ignore grossly incorrect training labels (Fig. 10e-f) suggests that the Error Learner has successfully decoupled the physical feature learning from the noisy supervision. This highlights LaNCoR's ability to mitigate the impact of label noise through feature alignment, prioritizing the intrinsic signal structure over unreliable annotations. While standard models may exhibit similar robustness in low-noise regimes, our experimental results (Table S1) indicate that they tend to overfit and memorize errors as noise levels increase. LaNCoR, however, maintains this ability to disregard erroneous labels and generalize to the true signal structure even under severe noise conditions. This resilience to noisy labels highlights LaNCoR's ability to mitigate their impact on model performance through feature alignment.

The remarkable stability of LaNCoR stems from several core design elements: an error learner, reparameterization, random Fourier features, and contrastive learning-based feature alignment. The self-attention mechanism-based error learner is vital for mitigating noisy labels. It corrects shifted label representation distributions by integrating the seismic wave feature space's representation distribution. To prevent negative impacts on seismic wave feature extraction from these shifts, gradient propagation in this module to the seismic wave representation is truncated. This capability of the error learner to effectively correct noise-induced label representation distribution shifts ensure the efficacy of feature alignment, as demonstrated by ablation experiments (Fig. 4).

Both reparameterization and random Fourier features contribute to multi-space feature representation alignment by introducing uncertainty. Reparameterization achieves this through variational inference, sampling from given prior distributions of feature representations. Conversely, random Fourier features align seismic wave feature representations and label representations in a lower-dimensional feature space using Gaussian kernels. Finally, feature alignment leverages contrastive learning principles to align feature and label representation distributions based on sample similarity.

LaNCoR is an effective, label noise-robust learning approach specifically designed for seismic signal processing. While its current application focuses on the picking task, utilizing symmetric seismogram-label mapping functions, its potential extends beyond this. By making minor adjustments to its label input and output, LaNCoR could be adapted for a wider range of tasks and integrated with various neural network architectures beyond SeisT and PhaseNet, offering a versatile robustness learning method.

However, it's important to acknowledge certain limitations. Balancing multiple optimization objectives is inherently complex. In this study, we relied on a heuristic search method. Although this approach might only achieve locally optimal solutions, the resulting loss balancing coefficients proved acceptable, as evidenced by our experimental results. We also explored a random search algorithm, but it was computationally expensive and produced fewer stable outcomes, failing to match the performance of the greedy search. Future research should prioritize developing more efficient and effective search techniques to derive more precise balance coefficients.

# Conclusion

This study introduces LaNCoR, a novel deep-learning model training approach for seismic signal processing, specifically applied to microseismic arrival time picking. LaNCoR tackles the problem of label noise, often introduced by human subjectivity, by harnessing the intrinsic consistency between waveform features and label distributions. The approach incorporates a feature alignment mechanism and an error learner to identify the correlation between microseismic waveform characteristics and label noise. This effectively reduces the influence of labeling inconsistencies stemming from human interpretation and environmental factors. Our experiments, conducted using real microseismic data from various coal mines, demonstrate LaNCoR's consistent performance across diverse noise levels and geographical regions. Comparisons with conventional training methods confirm LaNCoR's efficacy in improving model robustness and generalization, even when confronted with noisy label conditions.

# Acknowledgments

We would like to thank the anonymous reviewers for the review and insightful remarks.

# Declaration of Conflicting Interests

The authors declared no potential conflicts of interest regarding this article's research, authorship, and/ or publication.

# Funding

S.L. and S.M.M. were supported by NSF Award 2425714. X.Y. was supported by the National Natural Science Foundation of China under Grant 52504269.

# Data and Resources

The data used in this study cannot be publicly available.

# Code Available

Our source code is available at https://github.com/senli1073/LaNCoR and can reproduce the methods in this paper.